\documentclass[journal]{IEEEtran}
\usepackage{graphicx}
\usepackage{amsmath}

\usepackage{algpseudocode}%
\usepackage{url}
\let\labelindent\relax
\usepackage[titlenumbered,ruled]{algorithm2e}
\usepackage[inline]{enumitem}
\usepackage[colorinlistoftodos,prependcaption,textsize=tiny]{todonotes}
\usepackage{xspace} 
\usepackage{balance}
\newcommand{\td}[1]{}
\newcommand{\fs}[1]{}

\newcommand{\cs}[1]{}
\newcommand{\ad}[1]{}

\newcommand{\fzs}[1]{}
\newcommand{\adcomment}[1]{}

\usepackage{amssymb,amsmath}

\usepackage{setspace}
\usepackage{nopageno}
\begin{document}

\title{DxNAT - Deep Neural Networks for Explaining Non-Recurring Traffic Congestion}

\author{\IEEEauthorblockN{Fangzhou Sun, Abhishek Dubey, Jules White\\}
\IEEEauthorblockA{
Institute for Software Integrated Systems, Vanderbilt University \\ Nashville, TN, USA \\ \{fangzhou.sun,abhishek.dubey,jules.white\}@vanderbilt.edu}
}


%


\maketitle

\pagenumbering{arabic}
\thispagestyle{empty}

\begin{abstract}
Non-recurring traffic congestion is caused by temporary disruptions, such as accidents, sports games, adverse weather, etc. We use data related to real-time traffic speed, jam factors (a traffic congestion indicator), and events collected over a year from Nashville, TN to train a multi-layered deep neural network. The traffic dataset contains over 900 million data records. The network is thereafter used to classify the real-time data and identify anomalous operations. Compared with traditional approaches of using statistical or machine learning techniques, our model reaches an accuracy of 98.73 percent when identifying traffic congestion caused by football games. Our approach first encodes the traffic across a region as a scaled image. After that the image data from different timestamps is fused with event- and time-related data. Then a crossover operator is used as a data augmentation method to generate training datasets with more balanced classes.  Finally, we use the receiver operating characteristic (ROC) analysis to tune the sensitivity of the classifier. We present the analysis of the training time and the inference time separately.
\end{abstract}


%
\IEEEpeerreviewmaketitle

\section{Introduction}
\label{ch:trafficanomalydetection:sec:introduction}

\textbf{Emerging Trends.} Traffic congestion in urban areas has become a significant issue in recent years. Because of traffic congestion, people in the United States traveled an extra 6.9 billion hours and purchased an extra 3.1 billion gallons of fuel in 2014. The extra time and fuel cost were valued up to 160 billion dollars \cite{schrank20152015}. Congestion that is caused by accidents, roadwork, special events, or adverse weather is called non-recurring congestion (NRC) \cite{hall1993non}. Compared with the recurring congestion that happens repeatedly at particular times in the day, weekday and peak hours, NRC makes people unprepared and has a significant impact on urban mobility. For example, in the US, NRC accounts for two-thirds of the overall traffic delay in urban areas with a population of over one million \cite{lockwood200621st}. 

Driven by the concepts of the Internet of Things (IoT) and smart cities, various traffic sensors have been deployed in urban environments on a large scale. A number of techniques have been developed for knowledge discovery and data mining by integrating and utilizing the sensor data. Traffic data is widely available by using static sensors (e.g., loop detectors, radars, cameras, etc.) as well as mobile sensors (e.g., in-vehicle GPS and other crowdsensing techniques that use mobile phones). The fast development of sensor techniques enables the possibility of in-depth analysis of congestion and causes. 

The problem of finding anomalous traffic patterns is called traffic anomaly detection. Understanding and analyzing traffic anomalies, especially congestion patterns, is critical to helping city planners make better decisions to optimize urban transportation systems and reduce congestion conditions. To identify faulty sensors, many data-driven and model-driven methods have been proposed to incorporate historical and real-time data \cite{robinson2006development, lu2008faulty, zygouras2015towards, ghafouri2017optimal}. Some researchers \cite{kamijo2000traffic, veeraraghavan2005switching, yang2014detecting, kong2017lotad} have worked on detecting traffic events such as car accidents and congestion using videos, traffic, and vehicular ad hoc data. There are also researchers who have explored the root causes of anomalous traffic \cite{liu2011discovering, xu2013identifying, chow2014empirical, kwoczek2014predicting, kwoczek2015stuck, al2017distributed}. 

Most existing work still mainly focuses on a road section or a small network region to identify traffic congestion, but few studies explore non-recurring congestion and its causes for a large urban area. Recently, deep learning techniques have gained great success in many research fields (including image processing, speech recognition, bioinformatics, etc.), and provide a great opportunity to potentially solve the NRC identification and classification problem.  
There are still many open problems: (1) using feature vectors to represent traffic conditions loses the spatial information of the road segments, (2) using small and unbalanced dataset (traffic data with event labels is limited) to train neural networks downgrades the performance, a proper data augmentation mechanism is needed to balance the training data with different class labels, (3) building deep neural networks to model the traffic conditions of both recurring and non-recurring congestion.

\textbf{Contributions.}
In this paper, we propose DxNAT, a deep neural network model to identify non-recurring traffic congestion and explain its causes. To the best of our knowledge, our work is one of the first efforts to utilize deep learning techniques to study traffic congestion patterns and explain non-recurring congestion using events. The main contributions of our research are summarized as follows:
\begin{itemize}
\item We present an algorithm to efficiently convert traffic data in Traffic Message Channel (TMC) format to images
\item We introduce a crossover operator as a data augmentation method for training class balancing.
\item A convolutional neural network (CNN) is proposed to identify non-recurring traffic anomalies that are caused by events.
\item We create three scenarios to evaluate the performance of the proposed model by using real-world data of three events types (football games, hockey games, and traffic incidents). 
\end{itemize}

\textbf{Paper Organization.} The remainder of this paper is organized as follows:
Section~\ref{ch:trafficanomalydetection:sec:relatedwork} compares our work with related work;
Section~\ref{ch:trafficanomalydetection:sec:dataandexample} presents the dataset and a motivating example that explores the impact of football games on traffic congestion;
Section~\ref{ch:trafficanomalydetection:sec:formulation} formulates the problem;
Section~\ref{ch:trafficanomalydetection:sec:systemmodel} presents the solution approach;
Section~\ref{ch:trafficanomalydetection:sec:experiments} evaluates the performance of our model;
Section~\ref{ch:trafficanomalydetection:sec:conclusion} gives concluding remarks;

\section{Related Work and Challenges}
\label{ch:trafficanomalydetection:sec:relatedwork}
This section presents an overview of the related work on traffic anomaly detection, which includes studies about faulty traffic sensor detection, traffic event detection, and congestion cause indication. Three key research challenges and our contributions for detecting NRC are discussed in the end.

\textbf{Faulty Traffic Sensor Detection.}
Robinson et al. \cite{robinson2006development} proposed an approach that used data from inductive loop detectors to estimate travel time on road segments. His approach included a data cleaning method to clean the collected traffic data.
Lu et al. \cite{lu2008faulty} reviewed previous work on faulty inductive loops data analysis. 
Widhalm et al. \cite{widhalm2011identifying} presented a traffic anomaly detection method that used Floating-Car Data (FCD) as an independent information source. They developed a non-linear regression model to fit the traffic sensor data and FCD data.
Zygouras et al. \cite{zygouras2015towards} proposed a method comparing correlations among nearby sensors to identify faulty sensor readings. Their system was based on MapReduce paradigm to work for crowdsourcing data.
Ghafouri et al. \cite{ghafouri2017optimal} presented a faulty traffic sensor detection model based on Gaussian Processes. Particularly, they provided an effective approach for computing the parameters of detectors to minimize the loss due to false-positive and false-negative errors.

\textbf{Event Detection Using Traffic Data.} Monitoring traffic flow at intersections is important in the traffic event detection research.
Kamijo et al. \cite{kamijo2000traffic} developed an algorithm based on spatiotemporal Markov random field (MRF) for processing traffic images and tracking vehicles at intersections. Using the timeseries observed behaviors of vehicles, a hidden Markov model for accident detection is then proposed.
Veeraraghavan et al. \cite{veeraraghavan2005switching} presented a multiple cue-based approach combined with a switching Kalman filter for detecting vehicle events such as turning, stopping and slow moving.
Terroso-Sáenz et al. \cite{terroso2012cooperative} presented an event-driven architecture (EDA) that used vehicular ad hoc network and external data sources like weather conditions to detect traffic congestions. 
Yang et al. \cite{yang2014detecting} proposed a coupled Bayesian RPCA (BRPCA) model for detecting traffic events that used multiple traffic data streams.
Kong et al. \cite{kong2017lotad} proposed LoTAD to explore anomalous regions with long-term poor traffic situations. To model the traffic condition, crowd-sourced bus data is grouped into spatiotemporal segments. The segments with high anomaly indexes were combined to get anomalous regions.
Wang et al. \cite{wang2017road} proposed a two-stage solution to detect road traffic anomalies: (1) a Collaborative Path Inference (CPI) model that performs path inference incorporating static and dynamic features into a Conditional Random Field (CRF); (2) a road Anomaly Test (RAT) model calculates the anomalous degree for each road segment.

\textbf{Congestion Cause Indication.}
Liu et al. \cite{liu2011discovering} studied both known (planned) and unknown (unplanned) events behaving differently from daily network traffics as anomalies, and proposed algorithms that construct outlier causality trees based on temporal and spatial properties of detected outliers.
Xu et al. \cite{xu2013identifying} introduced an approach to identify urban congestion patterns based on the data cube. They proposed a multi-dimensional data analysis method for data cube.
Chow et al. \cite{chow2014empirical} presented an automatic number plate recognition technology to analyze urban traffic congestions and introduced a linear regression model to indicate the causes of the congestions.
Kwoczek et al. \cite{kwoczek2015stuck} proposed an Artificial Neural Network (ANN) based classifier to detect the road segments affected by planned events.
Mallah et al. \cite{al2017distributed} evaluated the performance of machine learning techniques for classifying congestions into different causes.  

\subsection{Research Challenge 1: Representing Heterogeneous Traffic Data and Event Labels Using Multi-Dimensional Images}
\label{ch:anomalydetection:gap:imagepresentation}

A feature vector is an n-dimensional vector and is the most popular representation of data objects. Besides numerical values, feature vectors can also represent texts and images. However, feature vectors may not be the best solutions for representing traffic and corresponding event labels.

Traffic conditions are highly affected by different influencing factors \cite{kwon2006components}, such as incidents, sports games, road work, weather, etc. The events and their physical locations are used as the labels. But since feature vectors have fixed length, it is not practical to manually encode the labels to a specific fixed length feature vector. More importantly, in pattern recognition and machine learning, features matter the most. When converting an image to a feature vector, you can directly convert the two-dimensional pixels to a one-dimensional vector, or you can first take the histogram of the image and then construct a feature vector that has several comparison metrics, such as mean, standard deviation, etc. Both methods will lose some relative spatial information in the original images. 

In contrast to feature vectors, images can preserve the original spatial relations by locating points on different pixels and can integrate multiple data sources by simply adding layers. 
Kwoczek et al. \cite{kwoczek2015stuck} showed a factor representation that integrates multiple features like event and weather into different layers in a data cube. However, though they mentioned the idea as a possible future work, they did not present any concrete solution to it.
Ma et al. \cite{ma2017learning} proposed a CNN-based approach for traffic prediction. They represented the traffic speed and time using a time-space matrix. The problem with the time-space matrix is that the spatial information between segments is lost, which is important in detecting traffic patterns because nearby roads usually show similar or related patterns. Additionally, their model simply considered traffic data, but there are many other factors affecting the future traffic conditions. Thus representing heterogeneous traffic and corresponding event labels using images remains a research gap. 

One of the key differences between our proposed approach and the existing ones is that we are trying to visualize the wide area sensor data distribution as Traffic Condition Images (TCIs), so that we can use CNN and other deep learning techniques for analyses.

\subsection{Research Challenge 2: Training Deep Learning Models Using Limited Data Instances}
\label{ch:anomalydetection:gap:datacrossover}

The performance of deep learning techniques highly relies on the quality of training data instances. However, the collected urban data may not provide enough data for training because of the data sampling rates. For example, our proposed model first converts traffic data to Traffic Condition Images (TCIs) and then trains different models using these images. But the traffic data we obtain from HERE \cite{heretrafficapi} is requested every minute. So for a day that consists of 1440 minutes, we will only have 1440 traffic images, which are too few for effective training purposes. The availability issue of data instances becomes worse considering there is also limited label data. It remains a research challenge of getting more training data using the existing data.

Traditional ways of solving this problem are: (1) waiting and collecting until enough training data is collected, (2) manually labeling the data, (3) adding data sources, e.g., collecting more data from social media. Our solution uses the idea of crossover from the genetic algorithm. We assume that traffic conditions within a short time range are associated with the same events. So we can apply a crossover operator on the TCIs to generate more TCIs with the same event label.

\subsection{Research Challenge 3: Modeling Traffic Patterns of Non-Recurring Events}
\label{ch:anomalydetection:gap:modelevents}

The existing work on traffic event detection focused on analyzing traffic videos or traffic sensor data streams to detect events that are directly related to traffic, such as vehicle stopping, car accidents, and road congestion. But few studies explored the contextual non-recurrent events whose impacts are also highly associated with certain traffic patterns. 

Recently there has been an explosion in research of using deep neural networks. But still, few have applied deep learning on studying traffic patterns. Deep learning techniques have gained great success in research fields like image processing, speech recognition, bioinformatics, etc. Convolutional neural networks are similar to original neural networks but convolutional layers are added in the front of the model to learn patterns in the original images. If traffic and label data can be converted to images, then CNN can be employed to learn their labeled patterns. 
It is still a research gap of how to develop an effective and efficient deep learning network for identifying and classifying traffic patterns of non-recurring events. 

We formulate the problem of identifying the specific traffic patterns associated with events in Section~\ref{ch:trafficanomalydetection:sec:images}, and then present the details of our proposed approach that uses convolutional neural networks in Section~\ref{ch:trafficanomalydetection:sec:classification}.

\section{Data and Motivating Example}
\label{ch:trafficanomalydetection:sec:dataandexample}

This section first introduces the datasets that we have integrated into the system, and then describes a motivating example in which we use the collected datasets to study the impact of football games on the traffic congestion in the city.

\subsection{Datasets}

Since October 2016, we have been continuously collecting and storing real-time traffic data from HERE API \cite{heretrafficapi} for all major roads in the Nashville area. In order to explore the impact of contextual events on urban mobility, we also collect the data about incidents and sports games. We cooperate with the Nashville Fire Department \cite{nashvillefired} to access their incident datasets, and manually collect the information about sports games from the web. As illustrated in Table~\ref{ch:trafficanomalydetection:tbl:datasets}, the details of the datasets that we have integrated into the system are as follows:
\begin{itemize}
\item The traffic dataset provides the real-time traffic information on road segments, such as speed limit, real-time speed, jam factor (JF), etc. The dataset contains historical traffic data for 3049 TMC road segments in the Nashville area.
\item The sports game dataset contains the operation information about sports games, such as game type, start and end time, attendance, location, etc.
\item The incident dataset provides the detailed records of incidents and the responding vehicles. For each incident, it provides the coordinates, incident type, alert time, vehicle arrival and departure time, weather condition, etc.
\end{itemize}

\begin{table}[t]
 \caption{Data }
\centering
\begin{tabular}{|p{1.0cm}|p{2.1cm}|p{2.1cm}|p{2cm}|}
\hline 
       & Traffic           & Sports Game       & Accident          \\
\hline
Format & TMC               & JSON             & JSON              \\
\hline
Source & HERE API \cite{heretrafficapi}          & ESPN, hockey-reference.com               & Fire Department   \\
\hline
Update & Every Minute      & Manually          & Manually          \\
\hline
Size   & 155 GB            & 28 Games          & 387 MB            \\
\hline
Range  & 10/2016 - Present & 10/2016 - 12/2016 & 03/2014 - 03/2017 \\
\hline 
\end{tabular}

 \label{ch:trafficanomalydetection:tbl:datasets}
\end{table}

\subsection{Motivating Example}
\label{ch:trafficanomalydetection:sec:example}

\begin{table}
\caption{The information of the eight football games studied in the motivating example}
\centering
\begin{tabular}{|p{1.2cm}|p{1.2cm}|p{1.2cm}|p{1.4cm}|p{1.2cm}|}
\hline
Date & Start Time (CST) & Stadium & Attendance & Duration \\
\hline
1/1/17 & 12:00 PM & Nissan Stadium & 65205 & 3:11 \\
\hline
12/11/16 & 12:00 PM & Nissan Stadium & 68780 & 3:02 \\
\hline
11/13/16 & 12:00 PM & Nissan Stadium & 69116 & 3:36 \\
\hline
10/27/16 & 7:26 PM & Nissan Stadium & 61619 & 3:08 \\
\hline
10/23/16 & 12:02 PM & Nissan Stadium & 65470 & 3:21 \\
\hline
10/16/16 & 12:02 PM & Nissan Stadium & 60897 & 3:12 \\
\hline
9/25/16 & 12:02 PM & Nissan Stadium & 62370 & 3:03 \\
\hline
9/11/16 & 12:05 PM & Nissan Stadium & 63816 & 2:57 \\
\hline
\end{tabular}
\label{ch:anomalydetection:tbl:footballgames}
\end{table}

The motivation for our research comes from a brief experiment, in which we study the impact of football games on the traffic congestion in the city. 

\begin{figure*}[t]
    \makebox[\linewidth]{
        \includegraphics[width=1.0\linewidth]{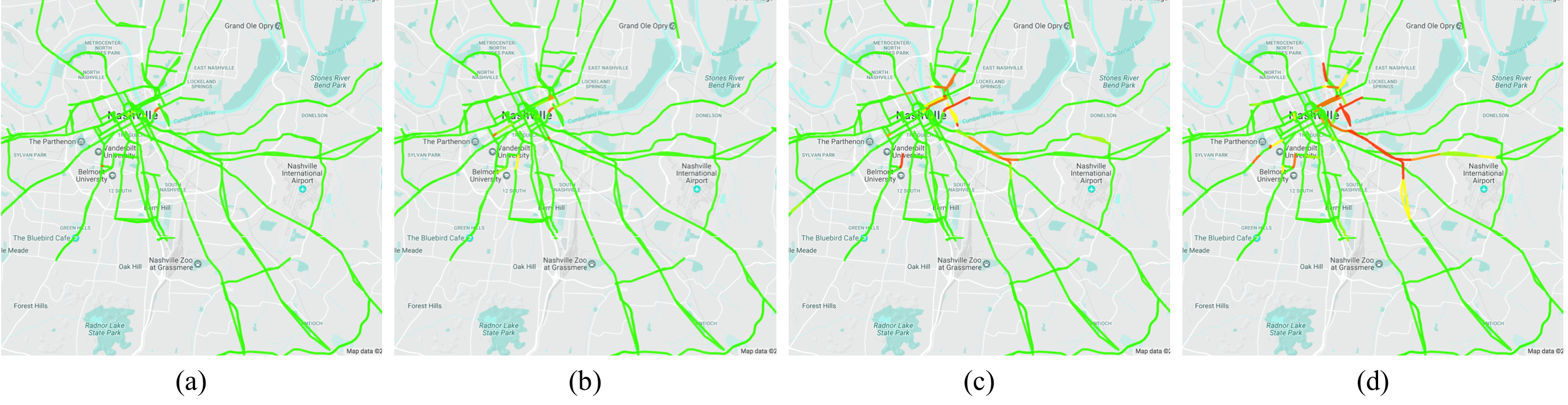}
    }
\caption{Impact of football games on traffic congestion in four one-hour time windows before football games: (a) from 4 hours to 3 hours, (b) from 3 hours to 2 hours, (c) from 2 hours to 1 hour, (d) from 1 hour to 0 hour.}
\label{ch:anomalydetection:fig:footballgametime}
\end{figure*}

During the studied period between Sept. 1, 2016 and Jan. 1, 2017, there were eight football games (as listed in Table~\ref{ch:anomalydetection:tbl:footballgames}) at the Nissan Stadium at downtown Nashville. During this time we collected data related to traffic (speed limit, real-time speed) and the football games (date, start time, duration, location)\footnote{Our dataset is larger. However, in this study we are focusing on these 4 months}.

To indicate the congestion condition, HERE \cite{heretrafficapi} provides a jam factor (JF) that ranges between 0.0 and 10.0 for each TMC road segment. In this study we compare the JF between the days when there is a football game and the days when there is no football game, during four one-hour time window directly before the games: $[-4, -3]$, $[-3, -2]$, $[-2, -1]$, and $[-1, 0]$ relative to the time when the game was scheduled\footnote{
Most football games are scheduled at 1 PM.
}.

As shown in Figure~\ref{ch:anomalydetection:fig:footballgametime}, the results of the JF difference on road segments in different time windows are visualized using heat maps. In the figure, colors ranging from green to red are used to indicate the small and big JF differences. The results show that the impact of football games on traffic congestion begins to increase from 4 hours before games.
We have observed this pattern across several game events in the city. Our hypothesis is that every event has a unique pattern and we can learn that pattern over time and use it to identify if a current congestion pattern matches with the expected pattern. If the pattern does not match then we can classify it as an anomaly.


\section{Problem Formulation}
\label{ch:trafficanomalydetection:sec:formulation}

In this section, we first provide a formal definition of the problem and then describe the assumptions for solving the problem.

\subsection{Definition}

The goal of this research is to model traffic patterns around the locations of non-recurrent events so that we can use the model to identify non-recurring congestion and its causes. 
The traffic pattern that we use here refers to the spatiotemporal relations of traffic speeds on many road segments in an area, which can be modeled and detected by a classifier. The definitions of all relative notions can be found in Table~\ref{ch:anomalydetection:tbl:notations}.

\begin{table}
\caption{Symbols used in the formulated problem}
\centering
\begin{tabular}{|p{0.17\columnwidth}|p{0.73\columnwidth}|}
\hline
$T$ & a timestamp \\
\hline
$t_{day}$ & time in the day in seconds \\
\hline
$t_{week}$ & weekdays encoded using integers (e.g., 0 for Sunday, 1 for Monday, etc.) \\
\hline
$t_{event}$ & time windows relative to events (e.g., 1 for the 1-hour time window before events) \\
\hline
$l_{event}$ & indicator of whether the current time is within a time window near the occurrence of an event \\
\hline
$S_e$ & a set of events in the city\\
\hline
$r$ & a road segment\\
\hline
$S_r$ & a set of road segments defined by TMC location codes \\
\hline
$S_t$ & a set of traffic data that contains speed limit and real-time speed for a set of road segments $S_r$ \\
\hline
$TMC_{key}$ & a string representing a road segment in $S_r$ \\
\hline
$TCI$ & traffic Condition Image, a gray-scale image to represent traffic conditions in a bounding box \\
\hline
$I_{w}$ &  the width of a TCI \\
\hline
$p$ & a pixel in TCI. Its value shows the normalized traffic speed on a road segment \\
\hline
$v_{r}$ & the real-time traffic speed (miles per hour) on a road segment $r$ \\
\hline
$TH$ & a threshold for the classifier to determine whether the input traffic data contains recurring or non-recurring congestion  \\
\hline
\end{tabular}
\label{ch:anomalydetection:tbl:notations}
\end{table}

The inputs to the system are data about traffic and events. Since the traffic data that we collected from HERE API is defined using Traffic Message Channel Location Code \cite{tmclocationcode} format (a standard for encoding geographic information), the road segments used in this study are also defined using the same TMC location codes. Event data is categorized with labels for training and validating purposes. The labels used are as follows:
\begin{itemize} 
\item Event-related: event indicator $l_{event}$, time window relative to the event $t_{event}$,
\item Time-related: time in the day $t_{day}$, weekday $t_{week}$
\end{itemize}

One of the key differences between our approach and the existing ones is that we are trying to visualize the wide area sensor data distribution as Traffic Condition Images (TCI), so that we can use CNN and other deep learning techniques to analyze and model the spatiotemporal relations. TCI is a $I_{w}$ by $I_{w}$ pixels image. Each pixel $p$ corresponds to a road segment in the real world and the grayscale value of each pixel represents the real-time traffic speed $v_r$ of the road segment $r$.

\textbf{Formulation of the Non-recurring Congestion Identification Problem.} 
Given a set of traffic data $S_t$ that contains speed limit and real-time speed for a set of road segments $S_r$ at a specific time $t_{day}$ on weekday $t_{week}$, and a set of event labels $S_e$, the model should determine $l_{event}$ that indicates whether the given traffic data contains congestion caused by a subset $S_e^{'}$ of event set $S_e$. If $l_{event}$ is true, the model should also provide the time window $t_{event}$ relative to events (i.e. $t_{event}$ can be used to estimate the event occurrence time).


Figure~\ref{ch:anomalydetection:fig:workflow} illustrates an example of the problem. Given raw traffic data at a specific time, the model should identify the possible non-recurring congestion and also provide an estimation of the event occurrence time.

\subsection{Assumptions}

The following assumptions are made when we design and formulate the non-recurring traffic congestion identification system: 
\begin{itemize}
\item We assume the availability of both traffic speed data and event information for the studied area and period.
\item We assume the traffic condition on a short road segment in a direction is the same anywhere on the segment.
\item We assume that an event happening in the urban environment will affect the traffic conditions of nearby road segments.
\item We assume that there is a robust correlation between the road segments affected by an event, and that the patterns can be identified by the image classification techniques of deep learning. 
\end{itemize}

\section{Our Approach}
\label{ch:trafficanomalydetection:sec:systemmodel}

\begin{figure*}[t]
\centering
\includegraphics[width=1.0\linewidth]{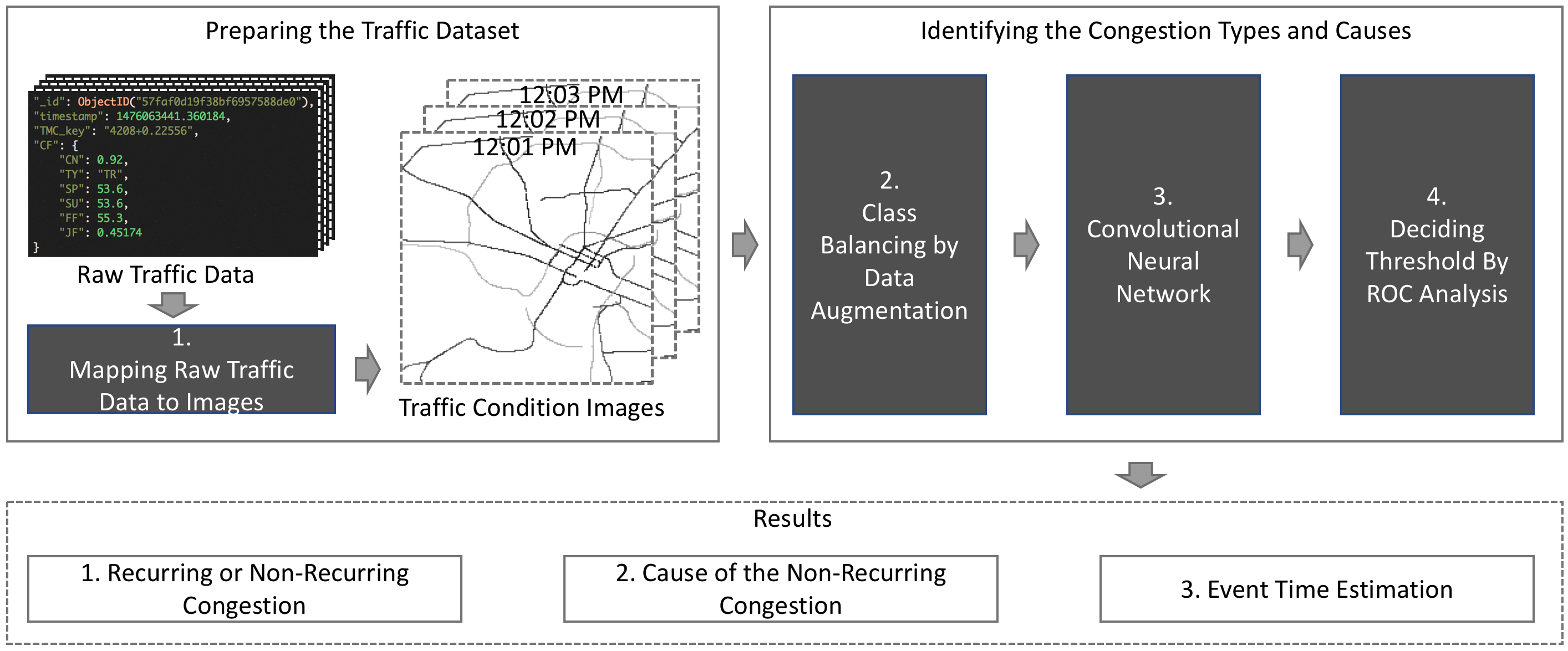}
\caption{Overall workflow of the non-recurring congestion identification system}
\label{ch:anomalydetection:fig:workflow}
\end{figure*}

In order to identify the specific traffic patterns associated with non-recurring events as defined in Section~\ref{ch:trafficanomalydetection:sec:formulation}, we present the details of our proposed approach in this section. The overall workflow of the system is shown in Figure~\ref{ch:anomalydetection:fig:workflow}. There are three key components in the system: (1) an algorithm that converts raw traffic data to images, (2) a convolutional neural network that classifies the traffic condition images, (3) ROC analysis that tunes the classification threshold to reduce the false positive and false negative rates.

\subsection{Feature Extraction by Mapping Traffic Data to Images}
\label{ch:trafficanomalydetection:sec:images}

Research challenge 1 describes the problem that feature vectors have limitations when representing urban data. To solve this issue, the first step is to convert the collected traffic data into images. We have been collecting real-time traffic data of Nashville area from Here Traffic API \cite{heretrafficapi} since Oct. 2016. The traffic data is encoded in TMC location codes. Since the TMC database is not open to the public, here we present an algorithm to convert traffic data for a specific time $T$ coded by TMC locations to traffic images. In order to project the traffic conditions to the pixels of images, we first initialize a gridded map and then re-sample the road segments defined by TMC location codes to the grids. The algorithm's input, output, and step details are as follows (for a set of road segments $S_r$, step 1 and 2 will run only once, but step 3 will run once for each timestamp):

\textbf{Input:} Traffic dataset $S_t$, road segment set $S_r$, and timestamp $T$. The raw traffic data of road segments $S_r$ for timestamp $T$ is queried from Traffic dataset $S_t$ in the database.

\textbf{Output:} A Traffic Condition Image (TCI).

\textbf{Step 1: Map grid initialization.} The map of the area containing the road segment set $S_r$ is divided into a map grid of squares. The length of each square is about 8.97 meters, so each grid cell covers about 80.51 square meters on the map. 

\textbf{Step 2: Road segment path re-sampling and smoothing.} The points from road segments are re-sampled to the centers of grid cells if the points are covered by the cell. Also, if the distance between two original points is large enough that there are blank cells between the two cells projected by the two points, then points will be interpolated to fill the blank cells. After this step, we get a two-dimensional array, in which each cell contains a list of TMC keys $TMC_{key}$ corresponding to points from road segments.

\textbf{Step 3: Traffic data projection to the images.} The two-dimensional array acts as a projecting table from original road segments to the image pixels. Now we can fill the images with traffic data by querying the traffic data using segment keys $TMC_{key}$ and timestamp $T$. We use the following equation to convert a traffic speed to a pixel value:

\begin{equation}
p =  
\begin{cases}
\frac{(80 - v_{s}^{t})*255}{80}, & \text{if}\ 0 \leqslant v_{s}^{t} \leqslant 80 \\
0, & \text{otherwise}
\end{cases}
\label{equa:speed-pixel}
\end{equation}
where $p$ denotes the pixel value (0-255) and $v_{s}^{t}$ denotes the real-time speed (miles per hour).

After getting initial projected TCI, simple image processing techniques are used to resize TCI to the desired size ($I_w$ by $I_w$).

\subsection{Data Augmentation by Crossover Operations}

TCI is our image representation of traffic speeds on road segments. Since traffic data is collected every minute, without data augmentation we can only get 1440 (the number of minutes in a day) TCIs for one day, which is usually not enough for training deep learning image processing models. To address research challenge 2 (i.e., the lack of enough traffic data with labels), we create a crossover operator to generate more labeled traffic condition images for training deep learning models. Crossover is originally a genetic operator from genetic algorithms to vary the chromosomes of individuals from one generation to another. We are motivated by a similar idea and present the crossover operator for our system: 
\begin{enumerate}
\item \textit{Getting TCI candidates.} For a given timestamp $T$, instead of only getting one TCI for $T$, we generate $n$ TCIs for time range $[T-t, T]$ ($t$ denotes a time length to extend $T$, e.g., 3 minutes). While these TCIs are the same in image size, they differ in the pixel values because they correspond to traffic speeds at different times.
\item \textit{Generating new labeled TCIs.} While looping through the pixels in TCI, for each pixel row there is a probability $p_{m}$ that its values will mutate and randomly select a new row from the same corresponding pixels in other TCI candidates. After the second step, we get a new TCI. Because we assume traffic conditions within a small time range are caused by the same events, we can give the new TCI the same event label. 
\end{enumerate}

The crossover operator can be executed for many times to generate many new data instances. Through crossover, we not only have more labeled data, but also reduce the probability of over-fitting in the training phase.

\subsection{Classifying Non-Recurring Congestion}
\label{ch:trafficanomalydetection:sec:classification}

In the previous section, we have described an algorithm that converts raw traffic data to TCI. Since the inputs contain images, it makes sense to apply convolutional neural networks. This section introduces our CNN model to classify the TCI using event labels. CNN is a class of deep and feed-forward artificial neural networks that have shown great success in image analysis tasks. Here we apply CNN to our problem that assigns event and congestion labels to a TCI. 

\begin{figure}[t]
    \makebox[\linewidth]{
        \includegraphics[width=1.0\linewidth]{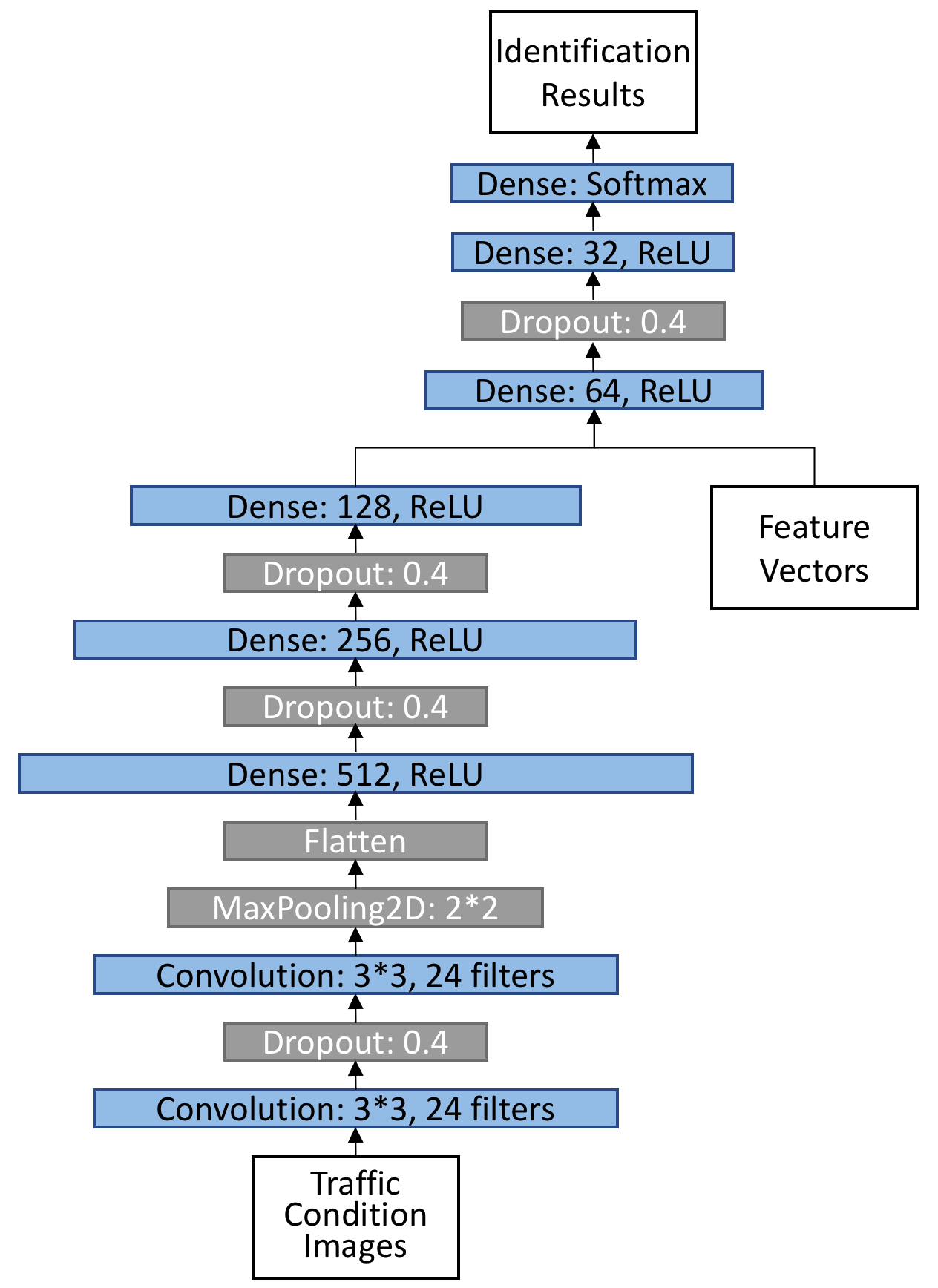}
    }
\caption{Our proposed convolutional neural network (CNN).}
\label{ch:trafficanomalydetection:fig:cnn}
\end{figure}

\textbf{CNN.} The architecture of the proposed CNN is shown in Figure~\ref{ch:trafficanomalydetection:fig:cnn}. Generally, the model consists of a stack of convolutional, fully-connected neural, dropout and max-pooling layers. Dropout layers are used throughout the model to prevent over fitting. Max pooling layers are used for spatial down-sampling. In the middle of the CNN, feature vectors that represent time of day and day of week that correspond to the TCI are concatenated to be input into the CNN to help it make better decisions. Details of the layer configuration, such as dimension, activation function, and dropout rate, can be found in the Figure~\ref{ch:trafficanomalydetection:fig:cnn}. Since we use one-hot encoding and the vectors are in categorical format (i.e., dimensional vector is all-zeros except for a one at the index corresponding to the class of the sample), categorical cross entropy is used as the loss function to train the model.

\begin{figure}[t]
    \makebox[\linewidth]{
        \includegraphics[width=1.0\linewidth]{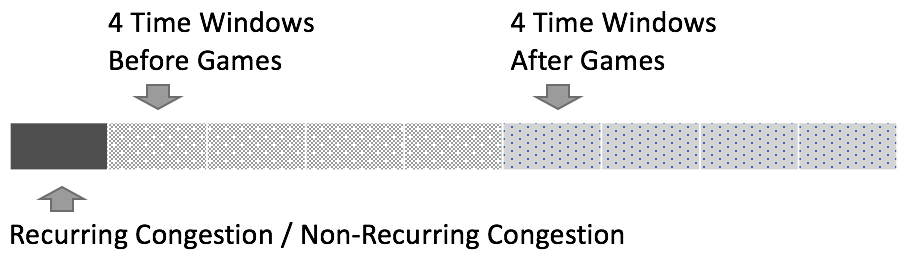}
    }
\caption{An example of the one-hot encoding format. Event labels are encoded using 9 classes: (1) first digit represents whether the traffic condition belongs to recurring congestion or non-recurring congestion, (2) if it's non-recurring congestion, the next 8 digits represent 8 time windows before and after events.
}
\label{ch:trafficanomalydetection:fig:onehot}
\end{figure}

\textbf{One-hot Encoding.} In the proposed CNN model, the input feature vectors are time in the day and weekday, and output labels are (1) whether the congestion in input TCI is recurring or non-recurring (2) the relative time windows that the TCI belongs to if it is non-recurring congestion. We use one-hot encoding to convert both input and output vectors to binary class matrix. The input matrix has 31 classes, in which 24 classes correspond to 24 hours and 7 classes correspond to 7 days of the week. As illustrated in Figure~\ref{ch:trafficanomalydetection:fig:onehot}, the output matrix has several classes, of which the first class represents whether the traffic condition belongs to recurring congestion or non-recurring congestion, and the next classes represent time windows before and after events. The first class is tunable since it directly determines whether the input traffic condition contains non-recurring congestion or not.
If the value of the first class output is higher than a predefined threshold $TH$, then the classifier will output that the input traffic data does not contain non-recurring congestion (even if the values of other classes are higher than the first class). The details of the tuning steps are presented in the following section.

\subsection{Tuning the Model Sensitivity by ROC Analysis}

Our approach uses receiver operating characteristic (ROC) analysis to tune the sensitivity of the CNN classifier.
ROC is a statistical plot that illustrates the diagnostic ability of a classifier system \cite{hanley1982meaning}. The ROC curve is a fundamental tool for diagnostic test evaluation. In an ROC curve, the true positive rate (TPR) is plotted in a function of the false positive rate (FPR). In machine learning, TPR represents sensitivity, recall or probability of detection, and FPR represents fall-out or false alarm \cite{roccurves}. By choosing a point from the curve, corresponding classification threshold can be decided.

In our model, the non-recurring congestion is considered as positive output and the recurring congestion is negative output.

\begin{figure}[tphb]
\begin{center}
\centerline{\includegraphics[width=1.0\columnwidth]{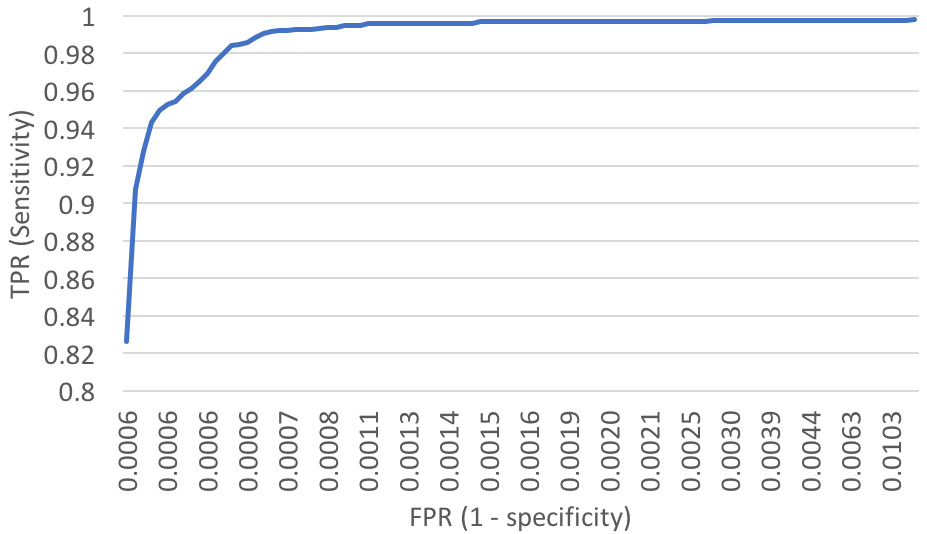}}
\caption{Receiver operating characteristics (ROC) curve analysis on the prediction threshold.}
\label{ch:trafficanomalydetection:fig:roc}
\end{center}
\end{figure}

We use the ROC analysis to tune the classification threshold that decides whether the traffic congestion in the input traffic data is recurring congestion or non-recurring congestion. We choose thresholds that range from 0.01 to 1.00 and the corresponding FPR and TPR of the training dataset are plotted (an example is shown in Figure~\ref{ch:trafficanomalydetection:fig:roc}). The curve's nearest point to the point (FPR: 0.0, TPR: 1.0) will be selected.

\section{Experiments}
\label{ch:trafficanomalydetection:sec:experiments}

In this section, we evaluate the proposed deep neural network's ability to identify non-recurring traffic anomalies by using real-world data of three event types: football games, hockey games, and traffic incidents. 
Keras \cite{keras} Python deep learning library is used to construct the models and TensorFlow \cite{abadi2016tensorflow} is selected as the tensor manipulation library.

\subsection{Scenarios}

\begin{figure}[tphb]
\begin{center}
\centerline{\includegraphics[width=1.0\columnwidth]{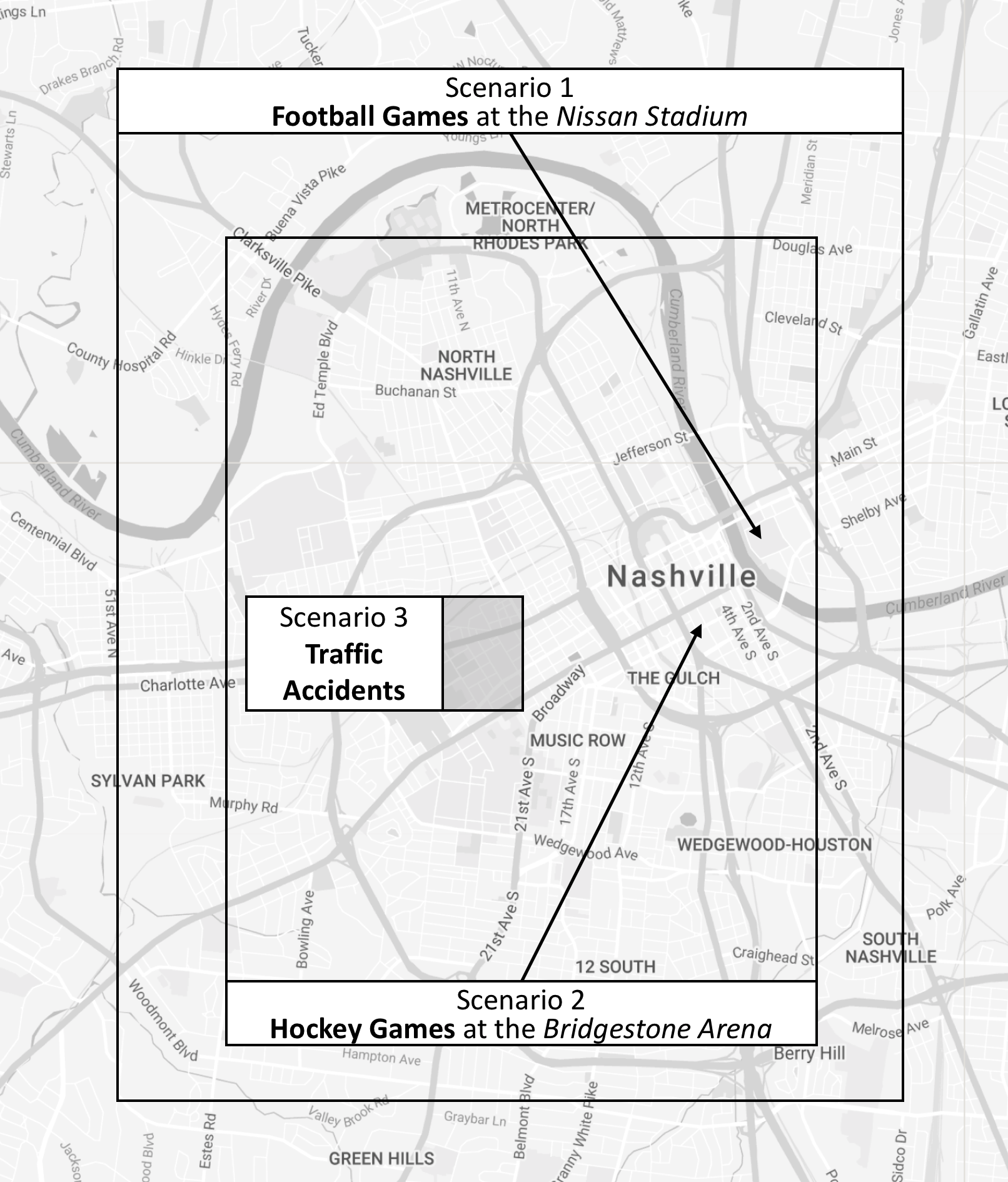}}
\caption{Experimental scenarios and their coverage areas: (1) detecting NRC caused by football games, (2) detecting NRC caused by hockey games, (3) detecting NRC caused by traffic accidents.}
\label{ch:trafficanomalydetection:fig:scenarios}
\end{center}
\end{figure}

As illustrated in Figure~\ref{ch:trafficanomalydetection:fig:scenarios}, we create three scenarios to test the performance of the proposed model. In each scenario, we consider one of the three event categories for training and validating the proposed model:
\begin{itemize}
\item \textit{Football Games.} Between Oct. 11, 2016, and Jan. 1, 2017, there were 8 NFL football games at the Nissan Stadium in Nashville. The traffic data in the bounding box (latitude range: [36.1120, 36.2052], longitude range: [-86.8475, -86.7543]) is used.
\item \textit{Hockey Games.} Between Oct. 14, 2016, and Jan. 03, 2017, there were 20 NHL hockey games at the Bridgestone Arena in Nashville. The traffic data in the bounding box (latitude range: [36.1237, 36.1936], longitude range: -86.8359, -86.7660]) is used.
\item \textit{Traffic Accidents.} Between Oct. 15, 2016, and Mar. 10, 2017, there were 23 traffic accidents at the selected block area. The traffic data in the bounding box (latitude range: [36.1470, 36.1586], longitude range: [-86.8126, -86.8009]) is used.
\end{itemize}

The traffic and event datasets are divided into two subsets for training and validation. The event and time information is encoded using one-hot encoding as described in Section~\ref{ch:trafficanomalydetection:sec:systemmodel}. An example of the output of our model is shown in Figure~\ref{ch:trafficanomalydetection:fig:onehot}. Particularly, the following metrics are used to define if an output is positive or negative: (1) an output is considered to be positive if it determines that the input TCI contains non-recurring congestion, (2) an output is negative if it determines the input TCI only contains recurring congestion.

\subsection{Experiment 1: Identifying NRC Caused by Football Games}

As shown in the motivating example in Section~\ref{ch:trafficanomalydetection:sec:example}, the 8 selected NFL football games have an average attendance of over 60,000 people, which shows a great impact on causing non-recurring traffic congestion. In the first scenario, we use the traffic data collected in 1-minute intervals between Oct. 11,
2016 and Jan. 1, 2017. Traffic data of 5 non-game days and two game days are used as the training dataset, and one non-game day and one game day are used as the validating dataset. As a comparison with the traditional machine learning techniques, we build a random forest model that uses the same training and validating dataset. Because random forests cannot use images directly as input, we first convert the traffic condition images to one-dimensional vectors, and then concentrate the traffic vectors with time of the day and day of the week vectors, and finally use the combined feature vector as input to the random forest model.

The accuracy, false positive rate (FPR) and false negative rate (FNR) of our model and the random forest model are shown in Table~\ref{ch:trafficanomalydetection:tbl:result1}. Our model outperforms the random forest model with higher accuracy and lower FPR and FNR.
\begin{table}[t]
\caption{Experiment results in scenario 1: training DxNAT for identifying NRC caused by football games}
\centering
\begin{tabular}{ |c|c|c|c| }
\hline 
 & Accuracy & FPR & FNR \\ 
\hline
DxNAT & 98.73\% & 1.57\% & 0.17\% \\ 
\hline
Random Forest & 84.06\% & 6.25\% & 2.17\% \\ 
\hline
\end{tabular}
\label{ch:trafficanomalydetection:tbl:result1}
\end{table}

\subsection{Experiment 2: Identifying NRC Caused by Hockey Games}

Compared with NFL football games, NHL hockey games in Nashville usually have less attendance (NHL 10,000 v.s. NFL 60,000). So we assume that an NHL hockey game has less impact on traffic conditions and it will be more difficult to detect the NRC related to hockey games.

In Scenario 2, we use traffic and hockey games data between Oct. 14, 2016, and Nov. 30, 2016, as the training dataset, and data of Dec. 15, 2016 (game day) and Dec. 16, 2016 (non-game day) as the validating dataset. The accuracy, FPR and FNR results are shown in Table~\ref{ch:trafficanomalydetection:tbl:result2}. Compared with the results in Scenario 1, the model has lower accuracy and higher FNR. Our assumption is validated that the NRC associated with hockey games with less attendance is harder to be detected.

\begin{table}[t]
\caption{Experiment results in scenario 2: training DxNAT for identifying NRC caused by hockey games}
\centering
\begin{tabular}{ |c|c|c|c| }
\hline 
 & Accuracy & FPR & FNR \\ 
\hline
DxNAT & 90.76\% & 8.11\% & 23.19\% \\ 
\hline
\end{tabular}

\label{ch:trafficanomalydetection:tbl:result2}
\end{table}

\subsection{Experiment 3: Identifying NRC Caused by Traffic Accidents}

In scenario 3, we explore the model's ability to detect NRC caused by road accidents. For the selected block area, there were eight traffic accidents on 7 different days between Oct. 18, 2016, and Dec. 13, 2016. We use six days with accidents as the training dataset and one day with an accident as the validating dataset. The DxNet model archives an accuracy of 86.59\% with FPR of 13.71\% and FNR of 4.44\% 

\begin{table}[t]
\caption{Experiment results in scenario 3: training DxNAT for identifying NRC caused by traffic accidents}
\centering
\begin{tabular}{ |c|c|c|c| }
\hline 
 & Accuracy & FPR & FNR \\ 
\hline
DxNAT & 86.59\% & 13.71\% & 4.44\% \\ 
\hline
\end{tabular}

\label{ch:trafficanomalydetection:tbl:result1}
\end{table}



\section{Conclusion}
\label{ch:trafficanomalydetection:sec:conclusion}

\begin{table}[t]
\caption{Summary of architectural decisions}
\centering
\begin{tabular}{ |p{3cm}|p{3cm}|p{1cm}| }
\hline 
Challenge & Approach & Section \\ 
\hline
Representing Heterogeneous Traffic Data and Event Labels &  Using Multi-dimensional Images & \ref{ch:anomalydetection:gap:imagepresentation} \\
\hline
Training Deep Learning Models Using Limited Data Instances & Developing crossover operator on original data
 & \ref{ch:anomalydetection:gap:datacrossover} \\
\hline
Modeling Traffic Patterns of Non-Recurring Events & Employing convolutional neural networks & \ref{ch:anomalydetection:gap:modelevents} \\
\hline
\end{tabular}

\label{ch:anomaly:tbl:summary-architectural}
\end{table}

In this paper, we propose a deep neural network model to identify non-recurring traffic congestion and explain its causes. To our best knowledge, our work is one of the first efforts to utilize deep learning techniques to study traffic congestion patterns and explain non-recurring congestion using events. Our main contributions are listed in Table~\ref{ch:anomaly:tbl:summary-architectural}. We present an algorithm to efficiently convert traffic data in Traffic Message Channel (TMC) format to images, as well as a crossover operator as a data augmentation mechanism for class balancing. A convolutional neural network is proposed to identify non-recurring traffic anomalies that are caused by events. We evaluate the proposed model by using three types of events (football games, hockey games, and traffic incidents). 

The future work to extend the current proposed model includes:
\begin{itemize}
\item \textit{Integrating more contextual features.} 
Existing work usually focuses just on traffic data, but there are many types of urban data available to help identify traffic patterns, like real-time bus travel time, speed, and weather.
Besides events, traffic conditions are affected by multiple environmental factors. The current work only considers time of day and day of week as the environmental training features. In the next step, we will include various features about weather conditions (such as humidity, nearest storm distance, visibility, etc.).
\item \textit{Identifying sizes of block areas and length of time windows.} For each event type (e.g., sports games, accidents), the size of impacting block areas as well as the number of impacting time windows in the experimental scenarios are selected arbitrarily. A mechanism is needed to automatically select the best impacting area size and time windows for each event type.
\end{itemize}

\section*{Acknowledgments}
This work is supported by The National Science Foundation under the award numbers CNS-1528799 and CNS-1647015 and a TIPS grant from Vanderbilt University.  We acknowledge the support provided by our partners from Nashville Metropolitan Transport Authority.

\balance
\bibliographystyle{IEEEtran}
\bibliography{papers}




%





\end{document}